 \documentclass[pmlr,twocolumn,10pt]{jmlr} 

\usepackage[utf8]{inputenc}
\usepackage{textgreek}
\usepackage{booktabs}
\usepackage{siunitx}
\usepackage{textcomp}
\usepackage{xparse}
\ExplSyntaxOn
\RenewDocumentCommand{\texttt}{m}
 {
  \tl_set:Nn \l_tmpa_tl { #1 }
  \tl_replace_all:Nnn \l_tmpa_tl { ' } { \textquotesingle }
  { \ttfamily \tl_use:N \l_tmpa_tl }
 }
\ExplSyntaxOff

\usepackage[svgnames]{xcolor}
\hypersetup{
    colorlinks=true,
    citecolor=NavyBlue,
    linkcolor=NavyBlue,
    urlcolor=NavyBlue,
}
\usepackage[nameinlink,capitalise,noabbrev]{cleveref}
\usepackage{caption}
\usepackage{soul}
\usepackage{xcolor} 

\usepackage[switch]{lineno}

\usepackage[table]{xcolor}
\definecolor{lightgray}{gray}{0.9}

\usepackage{listings}
\theorembodyfont{\upshape}
\theoremheaderfont{\scshape}
\theorempostheader{:}
\theoremsep{\newline}

\jmlrworkshop{Machine Learning for Health (ML4H) 2025} 

\title[An Adaptive ML Triage Framework for Predicting AD Progression]{An Adaptive Machine Learning Triage Framework for Predicting Alzheimer’s Disease Progression}

\author{%
 \Name{Richard Hou}$^1$ \Email{richard.hou@emory.edu} \\
 \Name{Shengpu Tang}$^{2,\dagger}$ \Email{shengpu.tang@emory.edu} \\
 \Name{Wei Jin}$^{2,\dagger}$ \Email{wei.jin@emory.edu} \\
 \addr \\[-1em]
 $^1$ Department of Biology, Emory University, USA \\
 $^2$ Department of Computer Science, Emory University, USA \\
 $^\dagger$ Equal contribution as senior authors.\\
}

\begin{document}

\maketitle

\begin{abstract}
Accurate predictions of conversion from mild cognitive impairment (MCI) to Alzheimer's disease (AD) can enable effective personalized therapy. While cognitive tests and clinical data are routinely collected, they lack the predictive power of PET scans and CSF biomarker analysis, which are prohibitively expensive to obtain for every patient. To address this cost-accuracy dilemma, we design a two-stage machine learning framework that selectively obtains advanced, costly features based on their predicted ``value of information''. We apply our framework to predict AD progression for MCI patients using data from the Alzheimer's Disease Neuroimaging Initiative (ADNI). Our framework reduces the need for advanced testing by 20\% while achieving a test AUROC of 0.929, comparable to the model that uses both basic and advanced features (AUROC=0.915, $p$=0.1010). We also provide an example interpretability analysis showing how one may explain the triage decision. Our work presents an interpretable, data-driven framework that optimizes AD diagnostic pathways and balances accuracy with cost, representing a step towards making early, reliable AD prediction more accessible in real-world practice. Future work should consider multiple categories of advanced features and larger-scale validation. 
\end{abstract}
\begin{keywords}
disease progression prediction, Alzheimer's disease, uncertainty estimation
\end{keywords}

\paragraph*{Data and Code Availability}
We used data from the Alzheimer's Disease Neuroimaging Initiative (ADNI)\footnotemark{} (\href{https://adni.loni.usc.edu/}{adni.loni.usc.edu}). The data is publicly available and can be requested on the ADNI website. Code for our experiments is available online.\footnotemark{}

\paragraph*{Institutional Review Board (IRB)}
This work does not require IRB approval. 
\footnotetext[1]{Data used in preparation of this article were obtained from the Alzheimer’s Disease Neuroimaging Initiative (ADNI) database. As such, the investigators within the ADNI contributed to the design and implementation of ADNI and/or provided data but did not participate in analysis or writing of this report.}
\footnotetext[2]{\url{https://github.com/chardhou-cpu/Triage-Framework-AD}}

\section{Introduction}
\label{sec:intro}
\vspace{-.5em}
Alzheimer's disease (AD), a major cause of dementia, is a neurodegenerative disorder marked by memory loss, cognitive decline, behavioral alterations, and diminished functional capabilities \citep{vaz_alzheimers_2020}. Over 40 million people worldwide currently suffer from dementia \citep{golde_alzheimers_2022}. A critical window for intervention lies in the stage of mild cognitive impairment (MCI), a transitional state where individuals exhibit cognitive deficits but maintain functional independence \citep{kelley_alzheimers_2007}. However, MCI is a heterogeneous condition; while many patients progress to AD, a significant portion remains stable or even reverts to normal cognition \citep{canevelli_spontaneous_2016}. Thus, identifying patients with MCI who are more likely to progress to AD is crucial to enable effective, personalized therapy \citep{golde_alzheimers_2018, li_predicting_2021}. 

Machine learning (ML) offers a promising solution to this problem by integrating complex, multimodal data to identify patterns that differentiate between progressive and stable MCI \citep{grueso_machine_2021}. However, the predictive power of ML models depends on the quality of the input features \citep{domingos_few_2012}. Biomarkers derived from positron emission tomography (PET) scans or cerebrospinal fluid (CSF) analysis are highly informative but can be too expensive and invasive for routine screening \citep{spasov_parameter-efficient_2019}. In contrast, low-cost features such as cognitive tests are more accessible but often lack diagnostic precision \citep{chen_prediction_2022,mitchell_can_2015,han_application_2017}. This trade-off between cost and accuracy limits reliable AD detection to specialized centers and impedes the development of scalable risk stratification approaches.


\begin{figure}[t]
    \centering
    \vspace{-1em}
    \includegraphics[width=\linewidth]{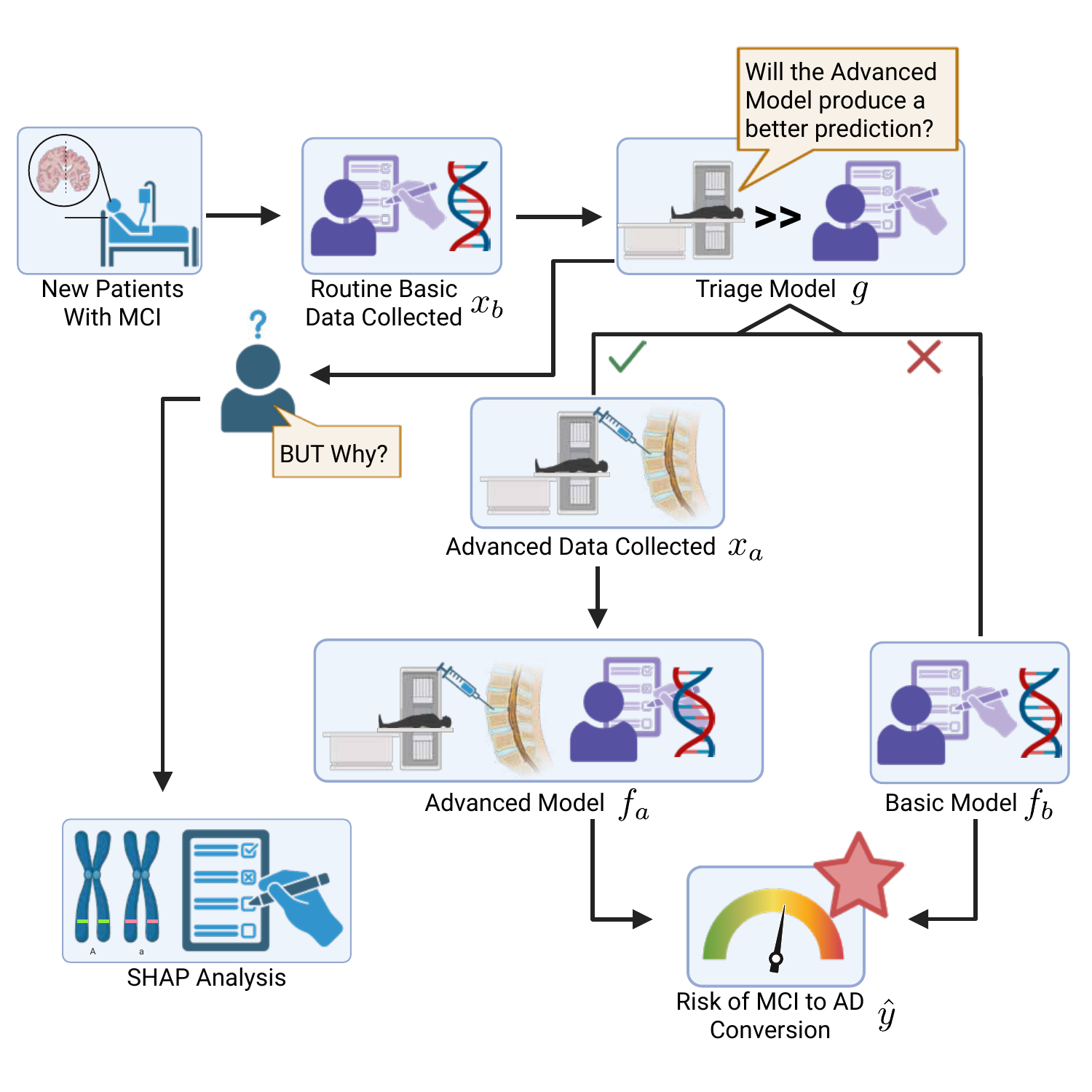}
    \vspace{-1em}
    \caption{Our proposed two-stage framework for AD progression prediction.  }
    \label{fig:workflow}
    \vspace{-1em}
\end{figure}


To address this challenge, we propose an \textbf{adaptive, two-stage} ML framework that mirrors how clinicians reason about initial screening and targeted follow-up testing (\cref{fig:workflow}). Our framework consists of three models: the \textbf{Basic Model}, the \textbf{Advanced Model}, and the \textbf{Triage Model}. The Basic Model leverages only low-cost, widely accessible clinical data and can be applied to all patients. The Advanced Model combines routine clinical data with biomarkers and PET scan data, and thus is only applicable to patients who have undergone these advanced testing. At the core of our framework is the Triage Model, which determines if a patient should undergo advanced testing. This decision is based on whether such testing is expected to \textit{improve diagnostic certainty}, ensuring that costly resources are used only where they are most likely to have an impact. In addition, the Triage Model is designed with interpretability in mind, providing clinicians with clear explanations of why escalation is recommended.


Our contributions are threefold: (1) We propose a cascading model that selectively allocates advanced tests for patients, which effectively reduces average diagnostic cost while preserving accuracy. (2) We develop an interpretable triage strategy that explicitly estimates when escalation is warranted, identifying cases in which advanced features are expected to provide meaningful diagnostic benefit. (3) We validate our framework on the ADNI dataset, demonstrating comparable AUROC to full-data models while lowering feature acquisition costs by up to 20\%. This work demonstrates how adaptive prediction can align machine learning with real-world clinical priorities, thereby advancing the development of cost-sensitive and interpretable systems for AD prediction.


\vspace{-1.em}
\section{Methods}
\label{sec:Methods}

\subsection{Method Overview}
\label{sec:overview}
Our framework consists of three model components (\cref{fig:workflow}): the Basic Model $f_b: \mathbb{R}^{d_b} \to [0,1]$, the Advanced Model $f_a: \mathbb{R}^{d_b + d_a} \to [0,1]$, and the Triage Model $g: \mathbb{R}^{d_b} \to [0,1]$. We use $d_b$ and $d_a$ to denote the dimensionality of the basic and advanced feature sets, respectively, with $x_b$ and $x_a$ denoting the corresponding feature vectors. 
We refer to the output $g(x)$ of the Triage Model as the escalation score. 
The final prediction $\hat{y}$ is determined as:
\begin{equation*} 
    \hat{y} = \begin{cases}
    f_b(x_b) & \text{if } g(x_b) \leq \tau, \\
    f_a(x_b, x_a) & \text{if } g(x_b) > \tau,
    \end{cases}
\end{equation*}
where the escalation threshold $\tau\in[0,1]$ is a hyperparameter that trades off coverage (fraction of patients handled by the Basic Model) and selective risk (error rate when not escalating).
In other words, the Basic Model $f_b$ is used by default; if escalation is deemed necessary according to the Triage Model, the Advanced Model $f_a$ is used. 
Let $p_b$ and $p_a$ denote the predicted probabilities for the positive class from the Basic and Advanced Models, respectively. We define the certainty of the model prediction as the absolute distance between the predicted probability and 0.5, $c = |p - 0.5|$. 
To train the Triage Model $g$, we use a binary supervision signal: $z = \mathbf{1}\left[\, c_a - c_b > \delta \,\right]$, where $\delta > 0$ is a margin parameter that specifies how much more certain the Advanced Model must be compared to the Basic Model to justify escalation. 
In this way, $z=1$ indicates that escalation provides a meaningful gain in certainty, while ${z}=0$ indicates that the Basic Model is likely already sufficiently certain. 
Thus, the Triage Model $g(x_b)$ is trained to anticipate when escalation to the Advanced Model is worthwhile.

\vskip 0.2em
\noindent\textbf{Interpretability of the Escalation Decision.} To make the escalation decisions transparent, we use SHAP \citep{lundberg_unified_2017}, a widely adopted method for feature attribution, to the Triage Model. 
For each patient, SHAP assigns a contribution value $\phi_j$ to feature $j$, representing the direction and magnitude of the feature's impact. 
Positive $\phi_j > 0$ indicates that larger values of a feature increase the likelihood of escalation, and vice versa.  In practice, we present the features with the strongest contributions together with their raw clinical values (e.g., MMSE score $= 27/30$ with $\phi = +0.81$).




\vspace{-1em}
\subsection{Implementation Details}
\vspace{-.5em}
\label{sec:experiments}

\noindent \textbf{Data Source and Cohort}. All data are obtained from the Alzheimer’s Disease Neuroimaging Initiative (ADNI) database (\href{https://adni.loni.usc.edu/}{adni.loni.usc.edu}). 
The primary goal of ADNI is to test whether serial MRI, PET, biological markers, and clinical assessments could be combined to measure the progression of MCI and early AD. Our final analysis include 1,142 participants who had been diagnosed with MCI.

\noindent \textbf{Feature Sets and Prediction Target.}
The Basic Model is trained using a basic feature set ($d_b=9$), which includes demographics (age, gender, race, education), APOE genetic status, and cognitive scores (MMSE, ADAS-11, Global CDR). The Advanced Model uses both the basic features and an advanced feature set ($d_a=329$) containing biomarker data from CSF analysis (Aβ\_42, t-tau, and p-tau levels) and PET imaging (amyloid status, hippocampal volume, etc). Advanced features are available for 551 participants. For both the Basic and Advanced Models, the prediction target is a binary indicator of whether a patient had converted from MCI to AD over a two-year period.
Finally, the Triage Model uses the same feature set as the Basic Model but is trained on the binary label defined in \cref{sec:overview}. 

\noindent \textbf{Data Partitioning.}
We carefully partition the available data to ensure robust evaluation. A held-out test set (n=100) is randomly sampled from the cohort (n=551) for which both basic and advanced features are available. The Basic Model is trained on the remaining cohort (n=1,042) who have basic features. The Advanced and Triage Models are trained on the remaining cohort (n=451) who have both basic and advanced features.

\noindent \textbf{Model Implementation.}
All models are built using scikit-learn pipelines that included preprocessing steps such as standardization of numerical features and one-hot encoding for categorical features, with 5-fold cross validation for hyperparameter and model selection. The final models are L2-regularized logistic regression classifiers for both the Basic Model and Advanced Model, and a support vector machine classifier with RBF kernel for the Triage Model, The final, optimized hyperparameters for each model are detailed in \cref{apd:first}. 
To prevent data leakage in training the Triage Model, its labels are generated using 5-fold cross-validation predictions of the Basic and Advanced Models.

\noindent \textbf{Evaluation and Baselines.}
We evaluate our framework on the held-out test set using standard binary classification metrics including AUROC, AUPRC, accuracy, precision, and recall. The escalation threshold for the Triage Model is selected by analyzing a risk-coverage curve on the training set. We compare our approach against several baselines with different heuristics, including the Basic Model, the Advanced Model, random escalation, and escalation based on the Basic Model's output probability or uncertainty. We also display a Cost-AUROC curve to illustrate the trade-off between diagnostic performance and the expected financial cost of testing.

\vspace{-1.em}
\section{Results}
\vspace{-0.5em}
\label{sec:results}
\noindent \textbf{Performance of Base Models.} 
The Advanced Model achieves an AUROC of 0.915 (95\% CI: 0.847-0.968) on the held-out test set, outperforming the Basic Model with an AUROC of 0.791 (95\% CI: 0.686-0.879) (\cref{tab:performance-comparison-ci}). While the basic features already achieves nontrivial performance, the advanced features provide a significant boost in predictive power. 

\begin{table*}[t]
    \centering
    \caption{Performance comparison with 95\% bootstrapped confidence intervals on the test set.}
    \vskip -1em
    \label{tab:performance-comparison-ci}
    \resizebox{\textwidth}{!}{%
    \begin{tabular}{lccccc}
        \toprule
        \bfseries Model & \bfseries AUROC (\%) & \bfseries AUPRC (\%) & \bfseries Accuracy (\%) & \bfseries Recall (\%) & \bfseries Precision (\%)\\
        \midrule
        Basic Model Alone & 79.1 (68.6 - 87.9) & 67.3 (48.8 - 81.8) & 72.0 (63.0 - 80.0) & 58.1 (40.0 - 74.1) & 54.5 (37.0 - 72.4) \\
        Advanced Model Alone & 91.5 (84.7 - 96.8) & 85.2 (72.1 - 94.2) & 82.0 (74.0 - 89.0) & 90.3 (78.1 - 100.0) & 65.1 (50.0 - 79.1) \\\midrule
        \rowcolor{lightgray} Our Triage Model (Thresh=0.05) & 92.8 (86.1 - 97.6) & 89.2 (78.9 - 95.9) & 83.0 (75.0 - 90.0) & 90.3 (78.1 - 100.0) & 66.7 (51.4 - 81.1) \\
        Baseline: Random 80\% Escalate & 89.0 (81.4 - 94.5) & 79.3 (64.0 - 90.7) & 80.0 (72.0 - 87.0) & 80.6 (66.7 - 93.3) & 64.1 (48.5 - 79.2) \\
        Baseline: Escalate 80\% Highest Prob & 91.7 (84.5 - 96.9) & 85.5 (72.4 - 94.4) & 83.0 (75.0 - 90.0) & 90.3 (78.1 - 100.0) & 66.7 (51.4 - 81.1) \\
        Baseline: Escalate 80\% Most Uncertain & 92.6 (85.7 - 97.4) & 88.7 (78.1 - 95.5) & 83.0 (75.0 - 90.0) & 90.3 (78.1 - 100.0) & 66.7 (51.4 - 81.1) \\
        \bottomrule
    \end{tabular}%
    \label{fig:Table}
    }
\end{table*}

\begin{figure}[b]
    \centering
    \vskip -.5em
    \includegraphics[width=0.75\linewidth,trim=0 0 0 34,clip]{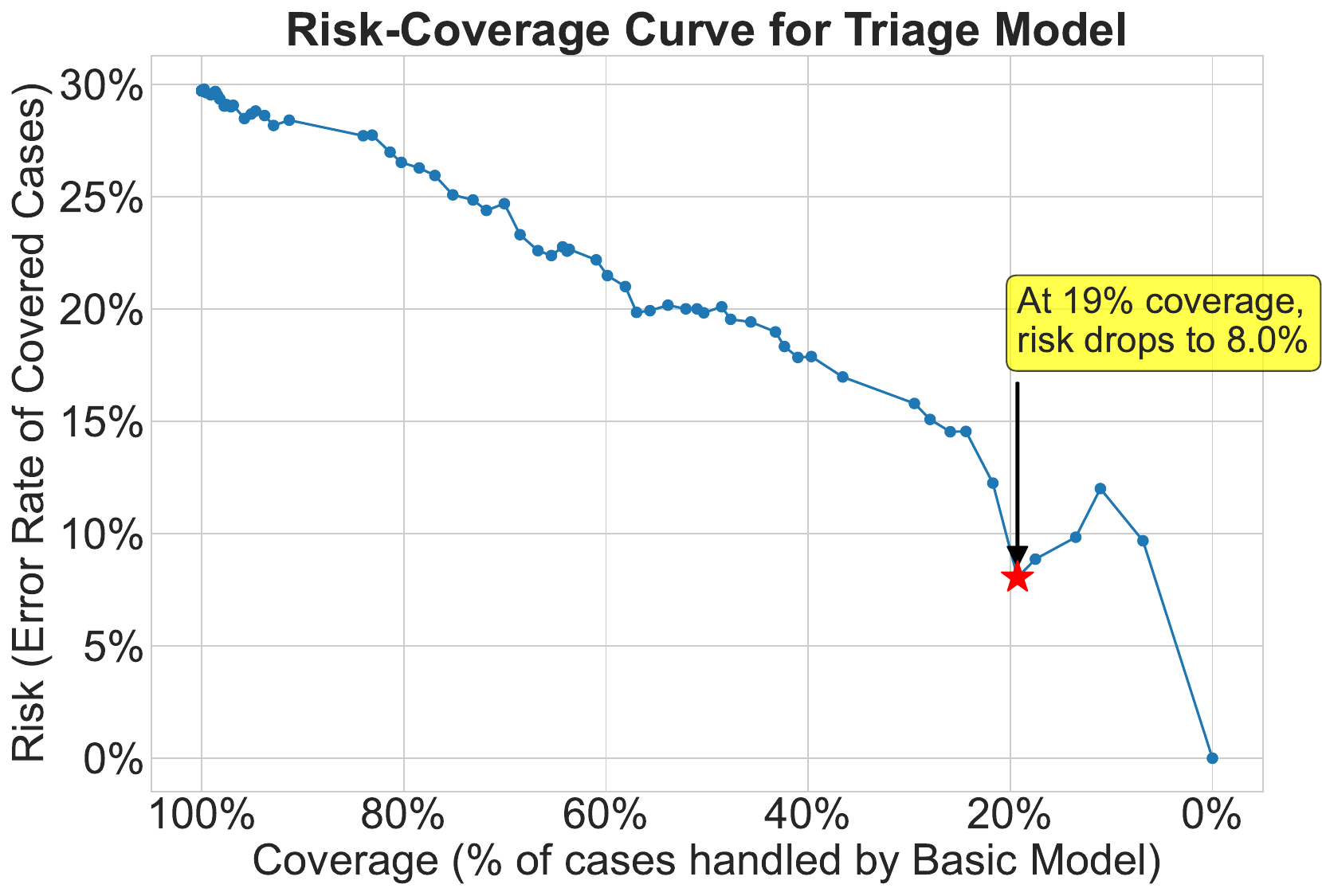}
    \vskip -.5em
    \caption{Risk-coverage curve and the selected threshold for the Triage Model. }
    \label{fig:Risk}
\end{figure}

\noindent \textbf{Triage Threshold Selection.} 
We set the certainty gain margin $\delta$ to be $0.2$, which created a reasonably balanced supervision signal for training the Triage Model while capturing a ``meaningful'' increase in prediction certainty. 
After hyperparameter tuning, the final Triage Model achieves a cross-validation AUROC of 0.77 in predicting the binary label representing the expected certainty gain. 
Based on the risk-coverage curve (\cref{fig:Risk}) that illustrates the trade-off between coverage (the percentage of patients handled by the Basic Model) and risk (the error rate for the non-escalated group), we observe an inflection point at $19\%$ coverage, where the error rate drops sharply to $8.0\%$. 
This corresponds to an escalation threshold of $\tau = 0.05$.

\noindent \textbf{Final Performance Comparison.} 
The performance of our proposed framework is summarized in \cref{fig:Table}, along with baselines that have the same escalation rate ($80\%$). 
Our proposed triage framework achieves an AUROC of 0.928 (95\% CI: 0.861-0.976) and an AUPRC of 0.892 (95\% CI: 0.789-0.959), outperforming the random escalation baseline and escalation based on Basic Model's probability predictions, and marginally surpassing escalation based on Basic Model's most uncertain cases. 
Notably, our triage framework achieves comparable performance to using the Advanced Model alone.
Based on the PR curves (\cref{fig:AUPRC}), our proposed framework can correctly identify nearly $60\%$ of positive cases (recall) while maintaining $100\%$ precision and $0$ false positives, whereas all other baselines can only achieve a recall of $< 30\%$ under the same condition.

\begin{figure*}[h]
    \centering
    \includegraphics[width=0.85\textwidth,trim=0 0 0 40]{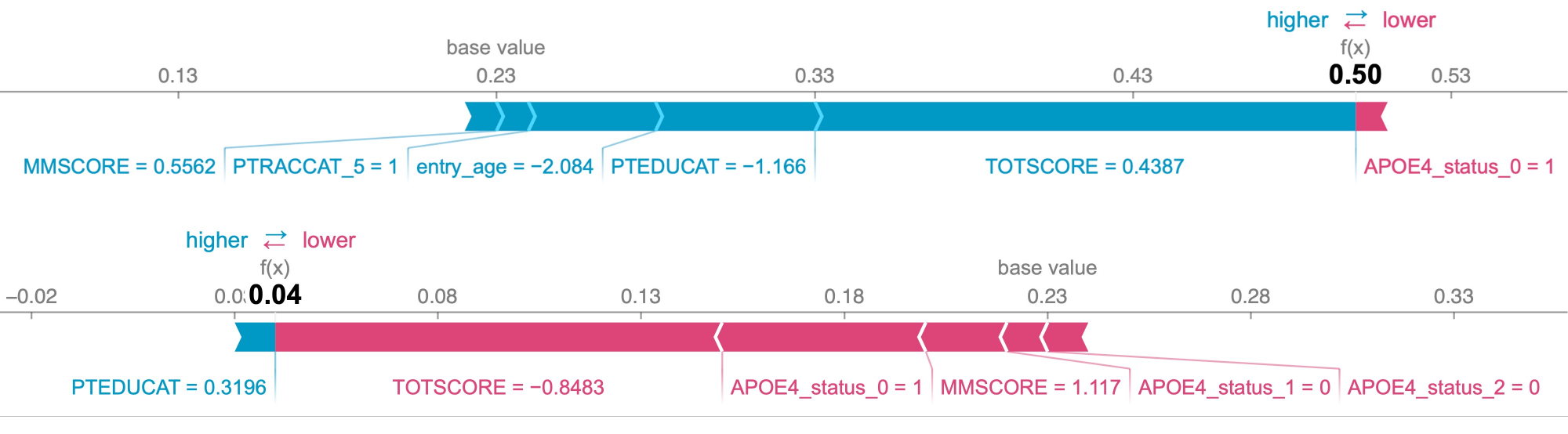}
    \vskip -1em
    \caption{SHAP force plots explaining the Triage Model's predictions recommending escalation (top) and recommending no escalation (bottom). {\small \textsf{TOTSCORE}} - normalized ADAS-11 total score; {\small \textsf{MMSCORE}} - normalized MMSE score; {\small \textsf{APOE4\_status}} - APOE4 carrier status. }
    \vskip -1em
    \label{fig:SHAP}
\end{figure*}

\noindent \textbf{Interpreting Triage Decisions.} 
We select two representative patients to illustrate how the Triage Model generates interpretable decisions. 
For the first case (\cref{fig:SHAP}-top), the Triage Model assigned a high escalation score of $0.50$, indicating a recommendation \textit{for} escalation to the Advanced Model. 
The most influential features include a high normalized ADAS-11 total score ($0.43$), a high normalized MMSE score ($0.56$), and the absence of the APOE4 allele. However, lower ADAS-11 and higher MMSE scores reflect good cognitive performance clinically \citep{cipolotti_neuropsychological_1995}, and the absence of APOE4 removes a major genetic risk factor for AD \citep{corder_gene_1993}. In this specific case, the high ADAS-11 score contradicts the high MMSE score and the absence of APOE4 allele. This conflicting profile justifies the need for an Advanced Test to confirm the diagnosis. 
For the second case (\cref{fig:SHAP}-bottom), the Triage Model assigned a low escalation score of $0.04$, indicating a recommendation \emph{against} escalation, with the most influential features being a low normalized ADAS-11 total score ($-0.85$), a high normalized MMSE score ($1.12$), and the absence of the APOE4 allele. Together, these characteristics form a coherent profile consistent with ``Stable MCI'', justifying the decision not to escalate.

\begin{figure}[h]
    \centering
    \includegraphics[width=0.8\linewidth,trim=0 0 0 34,clip]{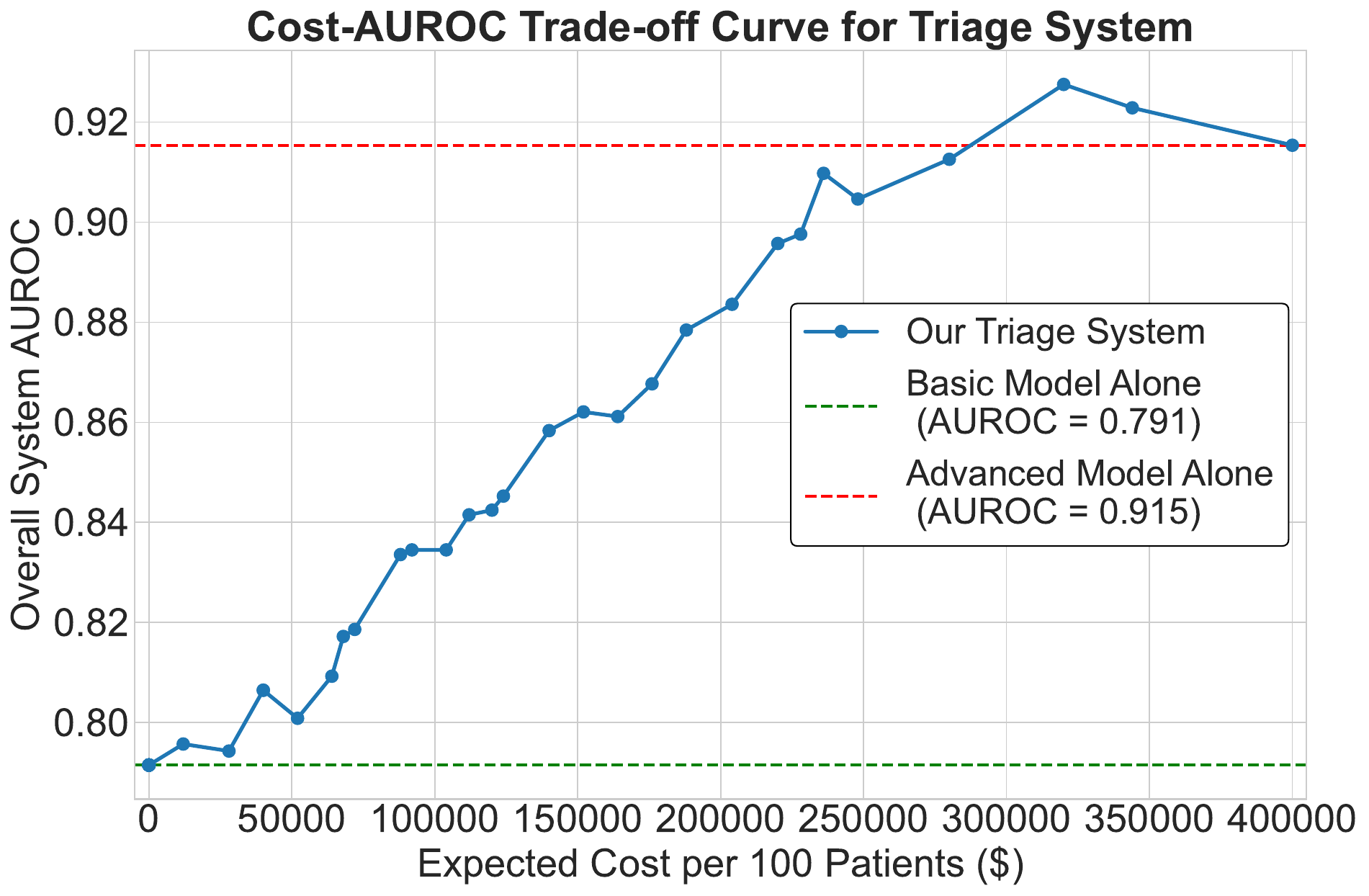}
    \vskip -.5em
    \caption{Cost-AUROC trade-off curve.}
    \vskip -1.em
    \label{fig:Cost}
\end{figure}

\noindent \textbf{Cost-Effectiveness Analysis.}
\cref{fig:Cost} shows the relationship between discriminative performance and the expected financial cost per 100 patients. 
For comparison, we also include the performance of the Basic Model alone (AUROC=0.79 at zero cost) and the Advanced Model alone (AUROC=0.92 at a maximum cost of \$400,000). 
In general, AUROC improves as the number of  escalations increases. Our selected operating point escalates approximately 80\% of patients, saving approximately \$80,000 per 100 patients \citep{alzheimers_association_2025_2025} compared to the Advanced Model while achieving a higher AUROC.

\vspace{-1.5em}
\section{Discussion \& Conclusion}
\vskip -.5em

In this work, we propose an adaptive two-stage ML framework for cost-effective prediction of Alzheimer's disease progression. Our framework integrates three models: a Basic Model trained on low-cost features, an Advanced Model that leverages additional costly features, and a Triage Model that selectively escalates patients when the additional features are expected to reduce model uncertainty. Empirically, our framework achieves performance comparable to the Advanced Model while potentially reducing resources required for obtaining expensive features. SHAP-based analysis further demonstrates that the triage decisions are clinically interpretable, with cognitive and genetic factors emerging as key drivers.

The potential real-world impact of our proposed framework is significant on both patients and the healthcare system. By accurately identifying 20\% of individuals with MCI who do not require escalation, our framework directly reduces patient burden. This could translate to 2.4 million individuals in the U.S. potentially avoiding invasive and costly procedures. The financial implications are equally significant, representing a potential societal cost saving of over \$9.6 billion \citep{alzheimers_association_2025_2025}. Therefore, our work demonstrates a powerful pathway to not only conserve clinical resources but, more importantly, to protect patients from unnecessary interventions.


Although our triage framework achieved statistically comparable performance to the uncertainty-based escalation baseline (\cref{tab:stat-comparison-scaled}), this does not mean that the Triage Model is redundant. Rather, it offers three distinct advantages: 

\vspace{-.5em}
\begin{enumerate}\setlength{\itemsep}{0pt}
\item Directness: The Triage Model is explicitly trained to predict whether performing an advanced test will yield a meaningful gain in diagnostic certainty. In contrast, the Basic Model's uncertainty only serves as a proxy; while correlated, it does not necessarily indicate that an advanced test will reduce uncertainty.

\item Interpretability: The dedicated Triage Model enables explanation of escalation decisions using techniques such as SHAP. These explanations (e.g., \cref{fig:SHAP}) are more transparent and informative than a generic statement that ``the Basic Model is uncertain''. 


\item Adaptability: While our paper considers a two-stage process with a binary escalation decision, the Triage Model naturally generalizes to a multi-class setting that could select among multiple advanced tests. Our approach thus lays important groundwork for devising more sophisticated, multi-step diagnostic protocols. 
\end{enumerate}

\vspace{-.5em}

\noindent \textbf{Limitations and Future Directions.} Due to the small dataset size, the training split was reused for Triage Model training and escalation threshold selection; future work with larger cohorts should use separate validation sets to ensure robust generalization. While we have defined the advanced feature set to include multiple data sources (CSF biomarkers and PET imaging), future work could consider separating these modalities to develop more granular, multi-step diagnostic pathways using techniques such as reinforcement learning \citep{tang2024towards}. Our framework uses static, cross-sectional data from a patient's baseline visit. Incorporating longitudinal data (e.g., time-series analysis of cognitive scores) would be a valuable extension and likely improve timeliness of diagnosis and actionability of the testing decisions \citep{tang2025transforming}. The interpretability of our Triage Model further enables clinician-in-the-loop decision making \citep{tang2020clinician}, where model explanations could support shared decision processes between human experts and AI. Future user studies involving clinicians could evaluate how such explanations affect trust, usability, and real-world adoption. Finally, our framework could be applied to other clinical tasks beyond predicting two-year AD progression. Future work could also compare our triage model against frontier approaches for producing escalation decisions, including large language models and AI agents \citep{xu_comprehensive_2025}.

\section*{Acknowledgments}
We thank the anonymous reviewers of ML4H 2025 for their valuable feedback. All content and code were verified by the authors, who take full responsibility for the results. This work was partially supported by the U.S. National Science Foundation under Award Numbers 2504088 and 2437345 (to W.J.). Any opinions, findings, and conclusions or recommendations expressed in this material are those of the authors and do not necessarily reflect the views of the National Science Foundation.









\clearpage

\appendix

\section{Additional Methods}\label{apd:first}

\noindent \textbf{ADNI Dataset.}
Data used in the preparation of this article were obtained from the Alzheimer’s Disease Neuroimaging Initiative (ADNI) database (\href{https://adni.loni.usc.edu/}{adni.loni.usc.edu}). The ADNI was launched in 2003 as a public-private partnership, led by Principal Investigator Michael W. Weiner, MD. The primary goal of ADNI has been to test whether serial magnetic resonance imaging (MRI), positron emission tomography (PET), other biological markers, and clinical and neuropsychological assessment can be combined to measure the progression of mild cognitive impairment (MCI) and early Alzheimer’s disease (AD).

\noindent \textbf{Cohort Demographics.}
In our experiments, we used random sampling without explicitly stratifying based on patient demographics. We have analyzed the demographic distributions of the training and test sets and found no statistical significant difference in any demographic dimension (\cref{tab:stat-comparison-patientdemo}).

\begin{table}[h]
    \centering
    \caption{Comparison of demographic characteristics between Training set and Test set.}
    \label{tab:stat-comparison-patientdemo}
    \resizebox{\columnwidth}{!}{%
    \begin{tabular}{l c c c}
\toprule
\textbf{Characteristic} & \textbf{Train Set (n=541)} & \textbf{Test Set (n=100)} & \textbf{p-value} \\
\midrule
\textbf{Age} & 71.5 $\pm$ 7.5 & 70.5 $\pm$ 7.1 & 0.23 \\
\textbf{Years of Education} & 16.1 $\pm$ 2.7 & 16.6 $\pm$ 2.4 & 0.12 \\
\textbf{Gender} & & & 0.63 \\
\hspace{1em} - Male & 247 (54.8\%) & 58 (58.0\%) & \\
\hspace{1em} - Female & 204 (45.2\%) & 42 (42.0\%) & \\
\textbf{Race} & & & 0.19 \\
\hspace{1em} - White & 423 (93.8\%) & 96 (96.0\%) & \\
\hspace{1em} - Black & 12 (2.7\%) & 1 (1.0\%) & \\
\hspace{1em} - Asian & 6 (1.3\%) & 0 (0.0\%) & \\
\hspace{1em} - Other & 10 (2.2\%) & 3 (1.0\%) & \\
\textbf{APOE4 status} & & & 0.58 \\
\hspace{1em} - 0 & 241 (53.4\%) & 55 (55.0\%) & \\
\hspace{1em} - 1 & 164 (36.4\%) & 32 (32.0\%) & \\
\hspace{1em} - 2 & 46 (10.2\%) & 13 (13.0\%) & \\
\textbf{APOE2 status} & & & 0.45 \\
\hspace{1em} - 0 & 416 (92.2\%) & 95 (95.0\%) & \\
\hspace{1em} - 1 & 35 (7.8\%) & 5 (5.0\%) & \\
\bottomrule
\end{tabular}%
    }
\end{table}

\noindent \textbf{Model Implementation Details.}
As mentioned in \cref{sec:experiments}, we report the final hyperparameters of the models below:

{\small \noindent
\textbf{Basic Model} \\
\hspace*{2em} \texttt{sklearn.linear\_model.LogisticRegression \\
\hspace*{2em} (C=0.1, penalty='l2', solver='liblinear')} \\
\textbf{Advanced Model} \\
\hspace*{2em} \texttt{sklearn.linear\_model.LogisticRegression \\
\hspace*{2em} (C=0.01, penalty='l2', solver='liblinear')} \\
\textbf{Triage Model} \\
\hspace*{2em} \texttt{sklearn.svm.SVC \\
\hspace*{2em} (C=10, kernel='rbf', gamma='auto')}
}




\section{Additional Results}\label{apd:second}
\begin{figure}[hbtp]
    \centering
    \includegraphics[width=0.85\linewidth,trim=0 0 0 34,clip]{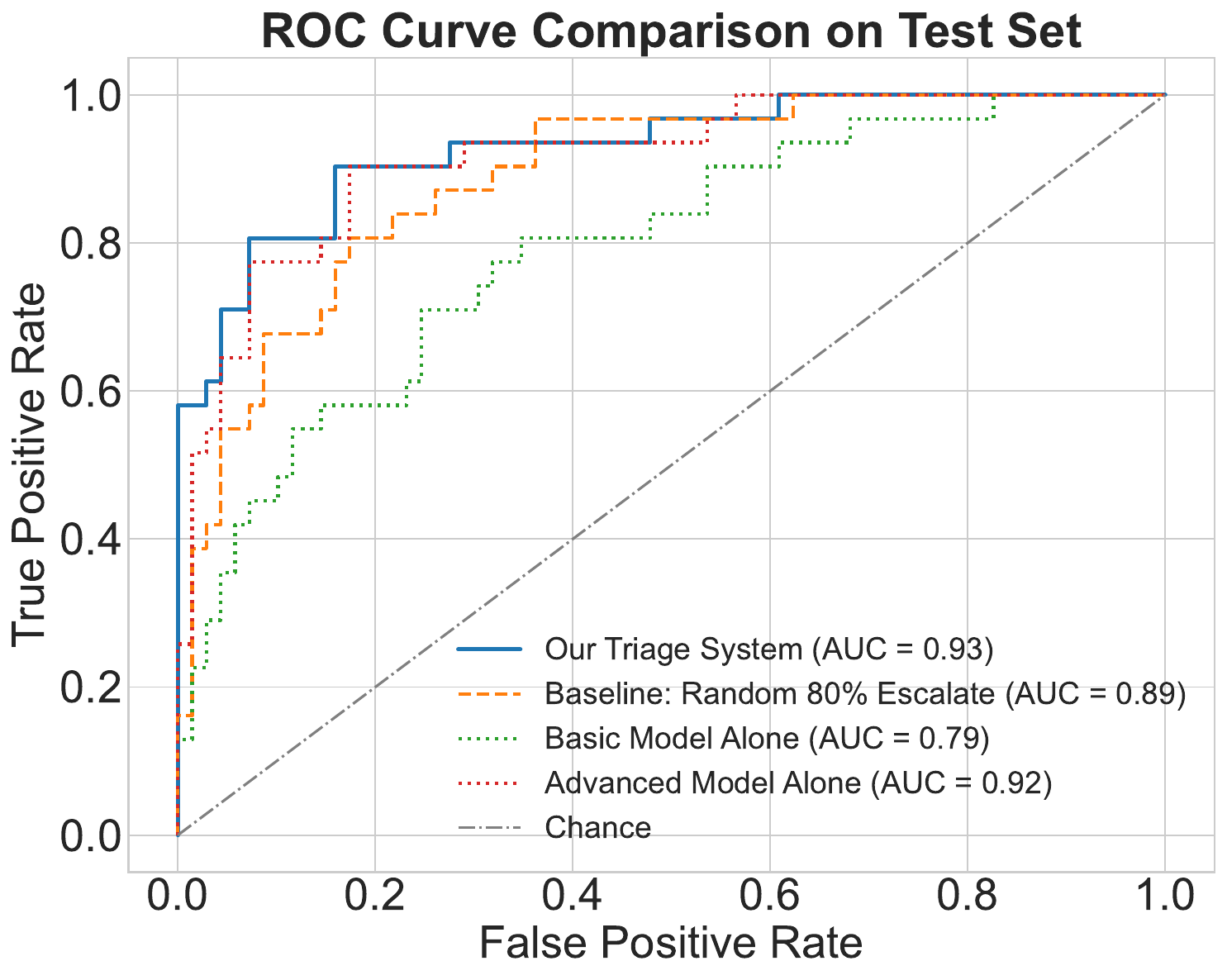}
    \caption{ROC curves for all models on the test set.}
    \label{fig:AUROC}
\end{figure}

\begin{figure}[hbtp]
    \centering
     \includegraphics[width=0.85\linewidth,trim=0 0 0 34,clip]{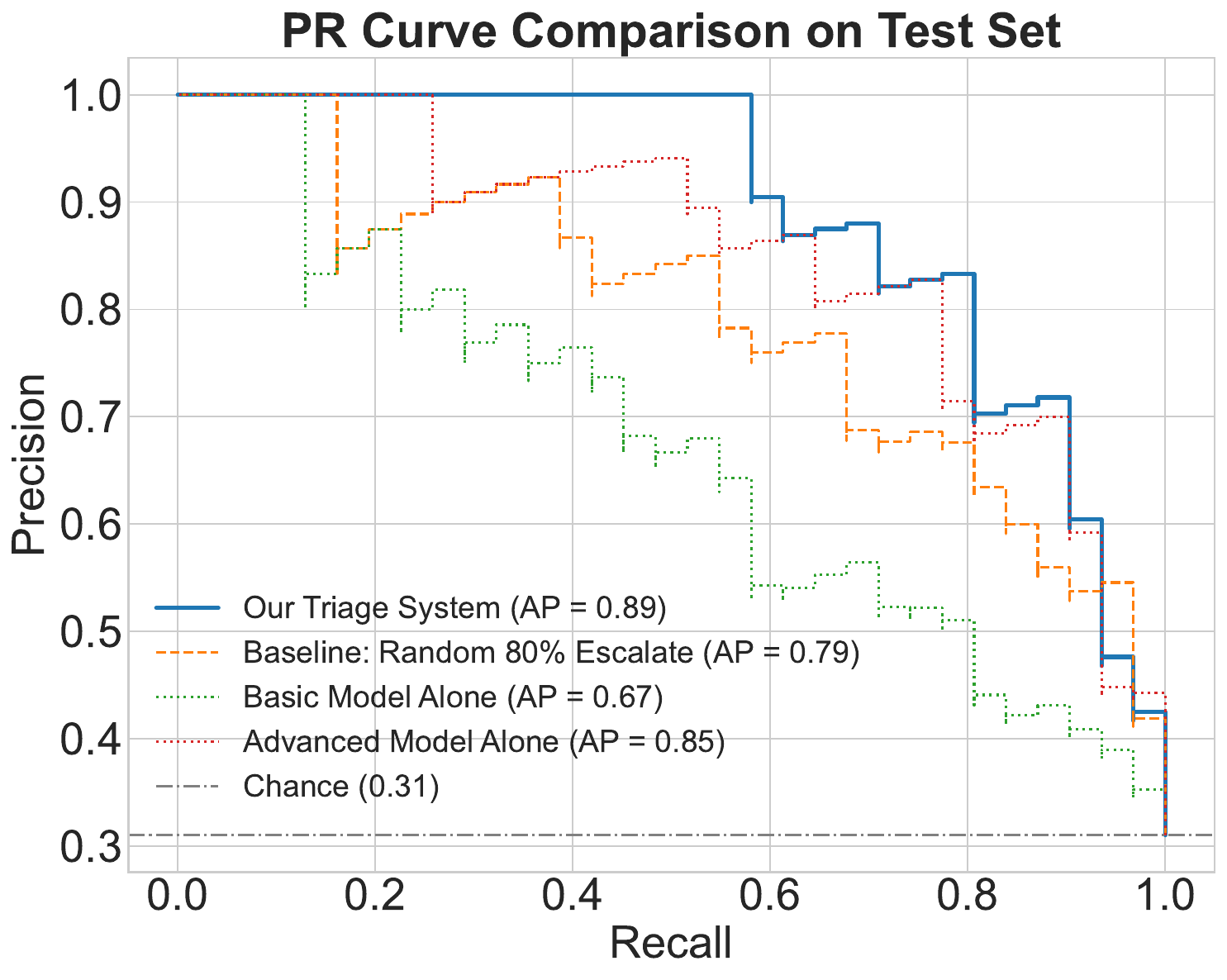}
    \caption{PR curves for all models on the test set.}
    \label{fig:AUPRC}
\end{figure}

\noindent \textbf{ROC and PR Curves.}
We further compare performance of our proposed approach against baselines by plotting ROC and PR curves on the test set. As shown in \cref{fig:AUROC}, the ROC curve of the Triage Model consistently arches higher and closer to the optimal top-left corner compared to all baselines (though it does not dominate all baselines). Similarly, the model’s PR curve also maintains a higher elevation across different recall values (\cref{fig:AUPRC}), demonstrating its ability to sustain generally higher precision than the baselines and single model. Notably, the PR curve reveals that our triage system maintains perfect precision up to a recall of nearly 60\%. This is clinically significant as it indicates that the top half of patients identified by the model as high-risk are classified with extremely high confidence, minimizing the risk of false positives in this critical subgroup.

\noindent \textbf{Statistical Significance of Performance Differences.}
We conducted bootstrapped resampling tests to compare the performance of our approach with the Basic Model alone, the Advanced Model alone, and several baselines, with AUROC differences detailed in \cref{tab:stat-comparison-scaled}. The Triage Model demonstrates a significantly higher AUROC than both the Basic Model alone (ΔAUROC = +0.1370, p=0.0010) and the baseline of randomly escalating 80\% of patients (ΔAUROC = +0.0388, p=0.0330). Although the Triage Model also numerically outperforms the more resource-intensive Advanced Model and the remaining baselines, these smaller performance gains were not statistically significant (p$>$0.05).
\begin{table} [h]
    \centering
    \caption{Statistical comparison of AUROC scores of the triage framework against baselines.}
    \label{tab:stat-comparison-scaled}
    \resizebox{\columnwidth}{!}{%
    \begin{tabular}{lccc}
        \toprule
        \bfseries Comparison Model & \bfseries $\Delta$AUROC & \bfseries p-value & \bfseries Significant? \\
        \midrule
        Basic Model Alone      & +0.1370 & 0.0010 & Yes \\
        Advanced Model Alone  & +0.0131 & 0.1010 & No  \\
        Baseline: Random 80 Escalate       & +0.0388 & 0.0330 & Yes \\
        Baseline: Escalate 80 Highest Prob   & +0.0112 & 0.0890 & No  \\
        Baseline: Escalate 80 Most Uncertain & +0.0028 & 0.5020 & No  \\
        \bottomrule
    \end{tabular}%
    }
\end{table}

\begin{figure}[hbtp]
    \centering
    \includegraphics[width=0.40\textwidth]{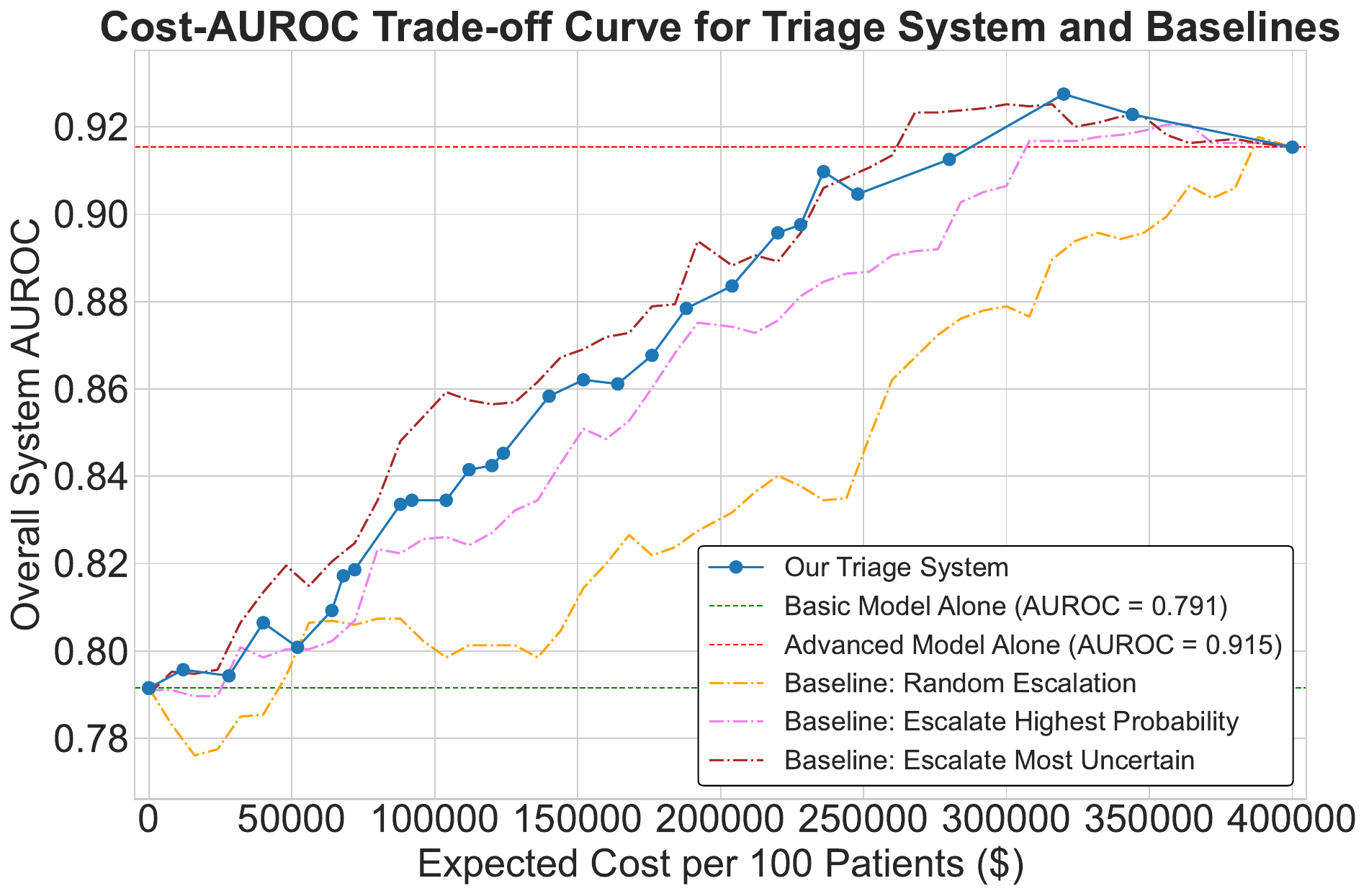}
    \caption{Cost-AUROC trade-off curves comparing the triage system with baselines.}
    \label{fig:Cost_all}
\end{figure}

\noindent \textbf{Comparative Cost-Effectiveness Analysis.}
We further compared various baselines in the cost-AUROC trade-off analysis (\cref{fig:Cost_all}). The triage system's performance trajectory (blue) remains consistently superior to Random Escalation baselines, but remained close to the other baselines, and it notably overlaps with the baseline that escalates based on the most uncertain cases.  
 
\noindent \textbf{Fairness Analysis.} To assess the fairness of our Triage Model, we analyzed the demographic distribution of the patient groups recommended for escalation versus those not escalated (at the $\tau=0.05$ threshold). We found no statistically significant difference in gender, race, or education level between the groups (\cref{tab:stat-comparison-patientdemo-escalation}). The one factor with a significant difference was APOE4 status (p=0.02), which is a primary clinical risk factor for AD. This suggests the model's triage decisions are driven by known clinical factors, not demographic bias. In future work, we will expand our study to a larger cohort and systematically evaluate and mitigate potential fairness issues \citep{xu_incorporating_2026}.

\begin{table}[h]
    \centering
    \caption{Comparison of demographic characteristics between escalated and non-escalated groups.}
    \label{tab:stat-comparison-patientdemo-escalation}
    \resizebox{\columnwidth}{!}{%
    \begin{tabular}{l c c c}
\toprule
\textbf{Characteristic} & \textbf{Escalated Group (\%)} & \textbf{Non-Escalated Group (\%)} & \textbf{p-value} \\
\midrule
\textbf{Education Level} & & & 0.93 \\
\hspace{1em} - 16+ years & 40\% & 42\% & \\
\hspace{1em} - 12-15 years & 35\% & 38\% & \\
\hspace{1em} - $<$ 12 years & 25\% & 20\% & \\
\textbf{Gender} & & & 0.77 \\
\hspace{1em} - Male & 55\% & 58\% & \\
\hspace{1em} - Female & 45\% & 42\% & \\
\textbf{Race} & & & 0.97 \\
\hspace{1em} - White & 85\% & 88\% & \\
\hspace{1em} - Black & 10\% & 8\% & \\
\hspace{1em} - Asian & 3\% & 2\% & \\
\hspace{1em} - Other & 2\% & 2\% & \\
\textbf{APOE4 Status} & & & 0.02 \\
\hspace{1em} - 0 & 30\% & 55\% & \\
\hspace{1em} - 1 & 45\% & 30\% & \\
\hspace{1em} - 2 & 25\% & 15\% & \\
\textbf{APOE2 Status} & & & 0.81 \\
\hspace{1em} - 0 & 10\% & 12\% & \\
\hspace{1em} - 1 & 90\% & 88\% & \\
\bottomrule
\end{tabular}%
    }
\end{table}

\end{document}